\documentclass[lettersize,journal]{IEEEtran}
\usepackage{amsmath,amsfonts}
\usepackage{algorithmic}
\usepackage{algorithm}
\usepackage{setspace}
\usepackage{array}
\usepackage{textcomp}
\usepackage{stfloats}
\usepackage{url}
\usepackage{verbatim}
\usepackage{graphicx}
\usepackage{subfigure}
\usepackage{subcaption}
\usepackage{cite}
\usepackage{booktabs}
\usepackage{multirow}
\usepackage{hyperref}
\hypersetup{hidelinks,
	colorlinks=true,
	allcolors=black,
	pdfstartview=Fit,
	breaklinks=true}

\def\BibTeX{{\rm B\kern-.05em{\sc i\kern-.025em b}\kern-.08em
    T\kern-.1667em\lower.7ex\hbox{E}\kern-.125emX}}
\usepackage{balance}

\begin{document}
\title{WaveRoRA: Wavelet Rotary Route Attention for Multivariate Time Series Forecasting}
\author{Aobo Liang, Yan Sun and Nadra Guizani
\thanks{A. Liang and Y. Sun are with the School of Computer Science (National Pilot Software Engineering School), Beijing University of Posts and Telecommunications, Beijing 100876, China. E-mail: liangaobo@bupt.edu.cn; sunyan@bupt.edu.cn.}
\thanks{N. Guizani is with the school of computer science and engineering, University of Texas at Arlington, Arlington, Texas, USA, 76019. E-mail: nadra.guizani@uta.edu.}
% <-this % stops a space
% \thanks{M. Guizani is with the Machine Learning Department, Mohamed Bin Zayed University of Artificial Intelligence (MBZUAI), Abu Dhabi, UAE. E-mail: mguizani@ieee.org.}
% \thanks{Y. Sun and M. Guizani are the corresponding authors}
}

\markboth{Journal of \LaTeX\ Class Files,~Vol.~18, No.~9, September~2020}%
{How to Use the IEEEtran \LaTeX \ Templates}

\maketitle

\begin{abstract}
In recent years, Transformer-based models (Transformers) have achieved significant success in multivariate time series forecasting (MTSF). However, previous works focus on extracting features either from the time domain or the frequency domain, which inadequately captures the trends and periodic characteristics. To address this issue, we propose a wavelet learning framework to model complex temporal dependencies of the time series data. The wavelet domain integrates both time and frequency information, allowing for the analysis of local characteristics of signals at different scales. Additionally, the Softmax self-attention mechanism used by Transformers has quadratic complexity, which leads to excessive computational costs when capturing long-term dependencies. Therefore, we propose a novel attention mechanism: Rotary Route Attention (RoRA). Unlike Softmax attention, RoRA utilizes rotary position embeddings to inject relative positional information to sequence tokens and introduces a small number of routing tokens $r$ to aggregate information from the $KV$ matrices and redistribute it to the $Q$ matrix, offering linear complexity. We further propose WaveRoRA, which leverages RoRA to capture inter-series dependencies in the wavelet domain. We conduct extensive experiments on eight real-world datasets. The results indicate that WaveRoRA outperforms existing state-of-the-art models while maintaining lower computational costs. Our code is available at \href{https://github.com/Leopold2333/WaveRoRA}{https://github.com/Leopold2333/WaveRoRA}.
\end{abstract}

\begin{IEEEkeywords}
Time series forecasting, Multivariate time series, Attention mechanism, Rotary position embeddings, Wavelet transform.
\end{IEEEkeywords}

\section{Introduction}
\IEEEPARstart{M}{ultivariate} time series (MTS) are prevalent in many real world scenarios, including weather\cite{zhang2022solar}, energy management\cite{uremovic2022new}, transportation\cite{guo2019attention} and network flow\cite{hewamalage2021recurrent}. Through analyzing historical observations, multivariate time series forecasting (MTSF) aims to extract the underlying patterns of time series data and provides future trends prediction. In the past few years, deep neural networks especially Transformer-based models (Transformers) have achieved great success in MTSF. Transformers rely on the Softmax self-attention (SA) mechanism\cite{vaswani2017attention} to effectively model long-term dependencies in sequences. Models such as Informer\cite{zhou2021informer} and PatchTST\cite{nie2023patch} make better predictions than previous works that are based on CNNs\cite{torres2021deep,wu2023timesnet} and RNNs\cite{liu2022msdr}. However, these approaches neglect the correlations between time series, thereby limiting their predictive performance. Some recent works have identified this issue. Models like Sageformer\cite{zhang2024sageformer} and iTransformer\cite{liu2023itransformer} focus on explicitly capturing inter-series dependencies and get superior performance. Nonetheless, SA has quadratic computational complexity and may hinder these models' scalability. Specifically, when modeling intra- or inter-series dependencies, SA incurs excessive computational costs as the sequence length or the number of variables increases, thus limiting its application in large-scale or high-dimensional scenarios\cite{jia2024witran, zhang2023crossformer}. This motivates us to develop a lightweight attention mechanism that simultaneously achieves the expressivity of of SA. In this paper, we design a novel \textbf{Ro}tary \textbf{R}oute \textbf{A}ttention mechanism (RoRA) for MTSF. We first introduce a set of routing tokens $r$ which has much fewer elements than that of the input sequence. The routing tokens aggregate information from the $KV$ matrices and redistribute it to the $Q$ matrix, offering linear complexity. Subsequently, we inject relative positional information through rotary positional embeddings (RoPE)\cite{su2024roformer} to enhance the feature diversity.

\begin{figure*}
    \centering
    \includegraphics[width=1.0\linewidth]{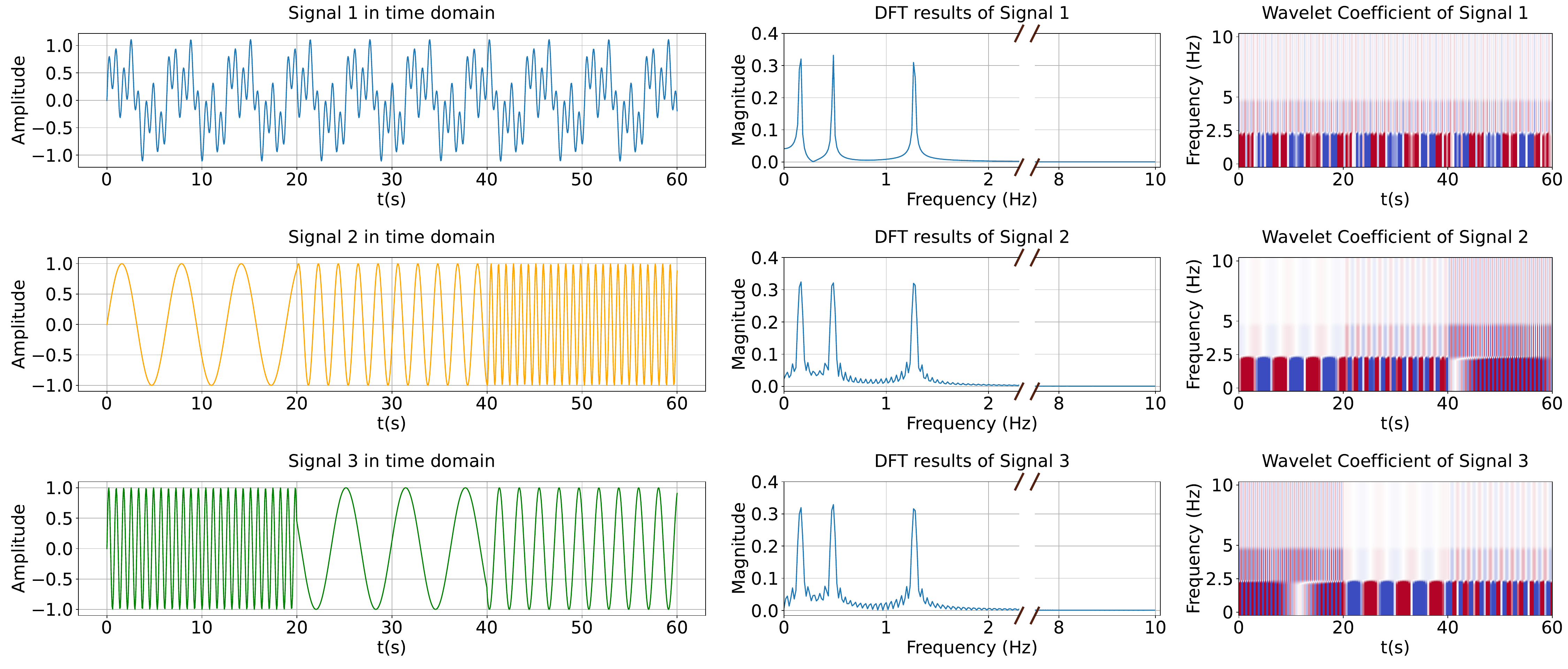}
    \caption{Three signals consisting of the same three functions: $\sin x$, $\sin 3x$ and $\sin 8x$ with the sampling rate 20Hz. Signal 1 is defined as $y=\sin x+\sin 3x+\sin 8x$, Signal 2 is evenly divided into 3 segments in the order of $\sin x$, $\sin 3x$ and $\sin 8x$, while signal 3 adjusts the order to $\sin 8x$, $\sin x$ and $\sin 3x$. These signals are different in the time domain but share similar patterns in the frequency domain. The rightmost column shows the wavelet coefficients obtained from a three-level DWT. Compared to the DFT results, the multi-layer wavelet coefficients preserve different periodic characteristics while better revealing the intervals where different periodic patterns are dominant.}
    \label{fig:1}
\end{figure*}

In addition, existing works mainly focus on analyzing the underlying patterns of MTS data in either time or frequency domains. Features of time domain and frequency domain can reflect different characteristics of the time series. For example, time domain features are conducive to extracting the trends of time series, while frequency domain features reveal the amplitude and phase information of each frequency component of the time series through discrete Fourier transform (DFT)\cite{cooley1965algorithm}. Compared with time domain features, frequency domain features intuitively reflect the periodicity or seasonality of time series\cite{liu2023temporal}. However, DFT faces the challenges in accurately fitting discontinuous signals, such as abrupt or non-triangular waveforms, which leads to the Gibbs phenomenon\cite{gibbs1899fourier,jiang2023fecam}. In addition, DFT has limitations in capturing the changes in the frequency domain over time. As shown in Figure \ref{fig:1}. Although the three signals exhibit  significant differences in the time domain, their frequency spectra are notably similar. Relying exclusively on frequency domain features may result in misleading information and negatively impact the predicting performance of the model. Recent works such as Autoformer\cite{wu2021autoformer} and FEDformer\cite{zhou2022fedformer} introduce frequency-domain features into time-domain representations to facilitate in extracting richer information. However, directly concatenating features of different domains may introduce noise to the model\cite{yang2023waveform}, resulting in suboptimal results.

% In addition, while Softmax attention demonstrates superior performance in capturing long-term dependencies, its quadratic computational complexity may hinder the scalability of the model. Specifically, when modeling intra- or inter-series dependencies, Softmax attention incurs excessive computational costs as the sequence length or the number of variables increases, thus limiting its application in large-scale or high-dimensional scenarios\cite{jia2024witran, zhang2023crossformer, liu2023itransformer}.

Therefore, our intuition is that extracting features in the wavelet domain could be beneficial for more accurate predictions. Wavelet transform inherits and develops the localization concept of short-time Fourier transform\cite{li2022short}, while overcoming its drawbacks such as the fixed window size that does not vary with different frequencies\cite{daubechies1990wavelet}. Wavelet transform extracts the time-frequency characteristics of a time series by scaling and translating a set of localized wavelet basis functions that decay within a short temporal window. 
For discrete time series, the Discrete Wavelet Transform (DWT)\cite{daubechies1992ten,cotter_2019} is typically employed, involving a multi-level wavelet decomposition. 
As illustrated in Fig. \ref{fig:1}, the wavelet coefficients produced from the multi-layer decomposition reflect the different periodic characteristics of the time series and accurately reveal the intervals where different periodic patterns are dominant. Our intuition is that extracting features in the wavelet domain allows us to simultaneously leverage the strengths of both time and frequency domains.

We decompose time series into multi-scale high-pass and low-pass components by DWT and design a multi-scale wavelet learning framework to facilitate the mutual conversion between wavelet coefficients and unified-size embeddings, aiming to capture temporal patterns within the time series. We then propose WaveRoRA to capture inter-series dependencies in the wavelet domain. Compared with SA, RoRA uses the relative positional information to facilitate in identifying the interactions between variables. Meanwhile, the routing tokens effectively alleviate the inherent redundancy between the attention weights and extract critical inter-series dependencies among different variables. Our contributions can be summarized as follows:
\begin{itemize}
    \item We propose modeling MTS data in the wavelet domain. To capture temporal patterns within the time series, we design a multi-scale wavelet learning framework that leverages both high-pass and low-pass components derived from DWT.
    \item We design a novel RoRA mechanism with linear complexity, introducing a small number of routing tokens $r$ to aggregate critical information from the $KV$ matrices and redistribute it to the $Q$ matrix. We then use RoPE to inject relative position information.
    \item Based on the wavelet learning framework and RoRA, we further propose WaveRoRA, which captures inter-series dependencies in the wavelet domain. We conduct extensive experiments on eight real-world datasets. WaveRoRA outperforms existing SOTA models while maintaining low computational costs.
\end{itemize}

\section{Related Work}
In this section, we summarize the typical methods of MTSF and provide a simple review of deep models combined with wavelet transform.
\subsection{Multivariate Time Series Forecasting}
MTSF aims to forecast future values based on a specific period of historical observations\cite{chen2023long}. Recently, Transformers achieve great success in MTSF with the Softmax self-attention mechanism\cite{vaswani2017attention}. However, the quadratic complexity of sequence length $L$ limits these models’ application on MTSF. Researches in recent years thus focus on balancing the predicting performance and computational efficiency of Transformers. LogTrans\cite{li2019enhancing} introduces a local convolutional module in self-attention layers and propose \textit{LogSparse} attention, which reduces the complexity of $O(L(\log L)^2)$. Informer\cite{zhou2021informer} proposes a \textit{ProbSparse} attention using a distilling technique and selecting top-k elements of the attention weight matrix to achieve $O(L\log L)$ complexity. Autoformer\cite{wu2021autoformer} designs a decomposition architecture with an \textit{Auto-Correlation} mechanism. It discovers the sub-series similarity based on the series periodicity and reduce the complexity to $O(L\log L)$. Despite improving forecasting performance and reducing computational costs, these models still rely on point-wise tokens, which limit the receptive field of the input sequence and may lead to overfitting in long-term forecasting. To address this issue, PatchTST\cite{nie2023patch} generates patch-wise tokens of each univariate series independently and capture intra-series dependencies. It reduces the computational complexity to $O((\frac{L}{P})^2)$, where $P$ denotes the patch length. The patch-wise tokens facilitates in expanding the receptive field of the input series, leading to better predicting performance. However, PatchTST overlooks the complex interactions between different variables. A recent model named iTransformer \cite{liu2023itransformer} generates series-wise tokens and inverts the attention layers to model the inter-series dependencies directly. Although iTransformer achieves superior performance, it still faces high computational costs of $O(M^2)$ when the number of variables $M$ is large.

\subsection{Wavelet Transform}
MTS data can be analyzed from various perspectives, such as the time domain, frequency domain, or wavelet domain, each offering unique insights of the underlying patterns of the time series\cite{zhang2022first}. Generally, the time domain and frequency domain provide intuitive features related to trends and periodic characteristics respectively\cite{bai2023crossfun}. Some existing works have explored combining features from these two domains. Autoformer\cite{wu2021autoformer} extracts series periods from frequency domain to assist its proposed \textit{Auto-correlation} in capturing dependencies between periodic segments. FEDformer\cite{zhou2022fedformer} randomly selects a fixed number of Fourier components to better represent a time series. Through combining Transformer with frequency analysis, FEDformer reduces the computational complexity while leading to performance improvement. However, the Fourier transform struggles to accurately fit discontinuous signals, such as abrupt changes or non-triangular waves. These types of signals are commonly found in real-world scenarios and can cause the Gibbs phenomenon\cite{gibbs1899fourier,jiang2023fecam}.

Analyzing time series in the wavelet domain naturally benefits from its ability to capture both temporal trends in the time domain and periodic patterns in the frequency domain. Wavelet transform decomposes a signal into different frequency and time scales by using scaled and shifted versions of a mother wavelet\cite{daubechies1992ten}. In recent years, several researches have been done to explore the effectiveness of modeling time series in the wavelet domain. TCDformer\cite{wan2024tcdformer} initially encoding abrupt changes using the local linear scaling approximation (LLSA) module and adopt Wavelet Attention to capture temporal dependencies. However, the model uses DWT to directly process point-wise tokens, which causes large computational costs. WaveForM\cite{yang2023waveform}, on the contrary, utilizes DWT to map the input series to latent representations in the wavelet domain. It uses dilated convolution and graph convolution modules to further capture the inter-series dependencies\cite{wu2020connecting}. However, WaveForM overlooks the correlation of wavelet coefficients at different levels, which may adversely affect the modeling of the periodic characteristics of time series data.

\section{Preliminaries}
In this section, we first introduce the concept of multivariate time series forecasting, then summarize the definition of DWT and several attention mechanisms.
\subsection{Multivariate Time Series Forecasting}
Given a multivariate historical observations $\mathbf{X}_{in}=\{\mathrm{x_1,x_2,...,}\mathrm{x}_L\}\in\mathbb{R}^{L\times M}$ with $L$ time steps and $M$ variables, MTSF aims to predict $\mathbf{X}_{out}=\{\mathrm{x}_{L+1},\mathrm{x}_{L+2}...,\mathrm{x}_{L+H}\}$, where $H$ denotes the prediction length.
\subsection{Discrete Wavelet Transform}
% DWT decomposes the signals multiple times to generate wavelets at different resolutions. It can capture the local characteristics of signals in both time and frequency domains, enabling localized analysis of temporal trends and frequency patterns. 

DWT decomposes a given signal $X$ multiple times to generate wavelet coefficients at different resolutions through a high-pass filter $\boldsymbol{h}$ and a low-pass filter $\boldsymbol{g}$. Typically, $\boldsymbol{h}$ and $\boldsymbol{g}$ correspond to a wavelet function $\psi(t)=a^{-\frac{1}{2}}\psi(\frac{t-b}{a})$ and a scaling function $\phi(t)=a^{-\frac{1}{2}}\phi(\frac{t-b}{a})$, which collectively form an orthogonal basis. The parameter $a$ is the scaling factor that controls the dilation of the wavelet while $b$ represents the translation factor that shifts the wavelet in time. Specifically, the process of DWT can be defined as Eq \ref{eq::1} and Eq \ref{eq::2}, where $j$ indicates the $j$-th decomposition, $y_h^{(j)},y_l^{(j)}$ are the high-pass and low-pass components and $y_l^{(0)}=X$, $S$ denotes the length of filters and $L^{(j)}=\lfloor\frac{L^{(j-1)}+S-1}{2}\rfloor$ refers to the length of $y_l^{(j)}$. $\boldsymbol{h}$ and $\boldsymbol{g}$ are performed convolution operations with the low-pass component of the former layer. After applying the convolutional filters, the frequency range of the signal is compressed to half of its original size. Therefore, the results can be downsampled by a factor of 2 according to the Nyquist theorem. It's worth noting that the scaling factor $a$ and translation factor $b$ of $\boldsymbol{h}$ and $\boldsymbol{g}$ require no adjustment because (a) the downsampling process simultaneously changes the resolution, allowing for multi-scale analysis; (b) the DWT process inherently encompasses translation as part of the wavelet decomposition.

% mother function $\psi(t)=a^{-\frac{1}{2}}\psi(\frac{t-b}{a})$ and a father function $\phi(t)=a^{-\frac{1}{2}}\phi(\frac{t-b}{a})$, extracting high-pass and low-pass results respectively at varying resolutions, where $a$ is the scaling factor that controls the dilation of the wavelet and $b$ is the translation factor that shifts the wavelet in time. Specifically, the process of DWT can be defined as Eq \ref{eq::1} and Eq \ref{eq::2}, where $j$ indicates the $j$-th decomposition, $yl^{(j)},yh^{(j)}$ represent the low-pass and high-pass results and $yl^{(0)}=X$, $S$ denotes the length of function $\psi$ and $\phi$ and $L^{(j)}=\lfloor\frac{L^{(j-1)}+S-1}{2}\rfloor$ is the length of $yl^{(j)}$. The high-pass and low-pass filters are performed convolution operations with the low-pass results of the former layer, generating temporary outputs $zh$ and $zl$. After the convolution, the results will be downsampled by a factor of 2 according to Nyquist theorem, denoted as $D_2$.

\begin{equation}\label{eq::1}
\begin{aligned}
y_h^{(j)}(n)&=\sum_{s=1}^{S}\boldsymbol{h}(s)y_l^{(j-1)}(2n-s),~n=1,2,...,L^{(j-1)}
\end{aligned}
\end{equation}

\begin{equation}\label{eq::2}
\begin{aligned}
y_l^{(j)}(n)&=\sum_{s=1}^{S}\boldsymbol{g}(s)y_l^{(j-1)}(2n-s),~n=1,2,...,L^{(j-1)}
\end{aligned}
\end{equation}

There will be $J+1$ wavelet coefficient components $\mathcal{C}=\{y_h^{(1)},y_h^{(2)}...,y_h^{(J)},y_l^{(J)}\}$ after $J$-level decomposition. Given a multivariate input $X\in\mathbb{R}^{L\times M}$, each output component contains the wavelet coefficient of each variable, denoted as $y_h^{(j)}=[y^{(j)}_{h,1},yh^{(j)}_2,...,yh^{(j)}_M]\in\mathbb{R}^{L^{(j)}\times M}$ and $yl^{(j)}=[yl^{(j)}_1,yl^{(j)}_2,...,yl^{(j)}_M]\in\mathbb{R}^{L^{(j)}\times M}$.

To reconstruct the time series, the Inverse Discrete Wavelet Transform (IDWT) uses the synthesis versions of $\boldsymbol{h}$ and $\boldsymbol{g}$, denoted as $\boldsymbol{h}^\prime$ and $\boldsymbol{g}^\prime$. The detailed process of IDWT can be defined in Eq \ref{eq::3}:
\begin{equation}\label{eq::3}
\begin{aligned}
\hat{y}_l^{(j-1)}(n)=&\sum_{s=1}^{S}\boldsymbol{h}^\prime(s)\tilde{y}_h^{(j)}(n-2s)+\\
&\sum_{s=1}^{S}\boldsymbol{g}^\prime(s)\tilde{y}_l^{(j)}(n-2s),\\
n=&1,2,...,2L^{(j)}
\end{aligned}
\end{equation}

\begin{equation}\label{eq::4}
\begin{aligned}
\tilde{y}_h^{(j)}(n)&=\begin{cases} 
y_h^{(j)}(\frac{n}{2}),  & \text{if }n\text{ is even} \\
0, & \text{if }n\text{ is odd}
\end{cases}\\
\tilde{y}_l^{(j)}(n)&=\begin{cases} 
\hat{y}_l^{(j)}(\frac{n}{2}),  & \text{if }n\text{ is even} \\
0, & \text{if }n\text{ is odd}
\end{cases}
\end{aligned}
\end{equation}
where $\tilde{y}_h^{(j)}$, $\tilde{y}_l^{(j)}$ are the upsampled results defined in Eq \ref{eq::4}, $\hat{y}_l^{(j)}$ is the reconstructed low-pass component of the $j$-th layer and $\hat{y}_l^{(J)}=y_l^{(J)}$. $\boldsymbol{h}^\prime$ and $\boldsymbol{g}^\prime$ are performed convolution operations respectively with the high-pass and low-pass components of the current layer. The output of $\hat{yl}^{(0)}\in\mathbb{R}^{L\times M}$ is the final reconstructed sequence in the time domain.

\subsection{Attention Mechanisms}
Given an input sequence $x\in\mathbb{R}^{N\times d}$ where $N$ denotes as the number of tokens and $d$ represents the token dimension, the attention mechanism can be defined as Eq \ref{eq::5}:
\begin{equation}
\label{eq::5}
\begin{aligned}
\mathrm{Attention}(Q_i,K_j,V_j)&=\sum_{j=1}^{N}\frac{\mathrm{Sim}(Q_i,K_j)}{\sum_{j=1}^N\mathrm{Sim}(Q_i,K_j)}V_j,\\
Q=xW_Q,K&=xW_K,V=xW_V
\end{aligned}
\end{equation}
where $W_Q,W_K,W_V\in\mathbb{R}^{d\times d}$ represent the projection matrices and $Q,K,V$ denotes query, key and value matrices respectively. 
The self-attention\cite{vaswani2017attention} uses $\mathrm{Sim}(Q_i,K_j)=\exp(Q_iK_j^T/\sqrt{d})$ to produce scaled $\mathrm{Softmax}$  attention (SA) scores. Such calculation processes lead to $O(N^2d)$ complexity.

To address the quadratic computational complexity, a variation of the self-attention mechanism is proposed, namely linear attention (LA)\cite{katharopoulos2020transformers}. linear attention is based on the assumption that there exists a kernel function $\Phi$ that can approximate the results of $\mathrm{Sim}(Q,K)$. Specifically, the function $\Phi$ aims to decouple the parameters of $\mathrm{Sim}(Q,K)$ and reformulate it as shown in Eq \ref{eq::6}:
\begin{equation}\label{eq::6}
\begin{aligned}
    \mathrm{LA}(Q_i,K_j,V_j)&=\sum_{j=1}^{N}\frac{\Phi(Q_i)\Phi(K_j)^T}{\sum_{j=1}^N\Phi(Q_i)\Phi(K_j)^T}V_j\\
    &=\frac{\Phi(Q_i)\sum_{j=1}^{N}\Phi(K_j)^TV_j}{\Phi(Q_i)\sum_{j=1}^{N}\Phi(K_j)^T},
\end{aligned}
\end{equation}
where $\Phi(x)=\mathrm{Elu}(x)+1$ typically. The order of matrix multiplication changes from $Q$-$K$ pairs to $K$-$V$ pairs, leading to $O(Nd^2)$ complexity. However, linear attention still faces the challenges with excessive computational complexity when dealing with high-dimensional input data.

\begin{figure*}[!htbp]
    \centering
    \includegraphics[width=1.0\linewidth]{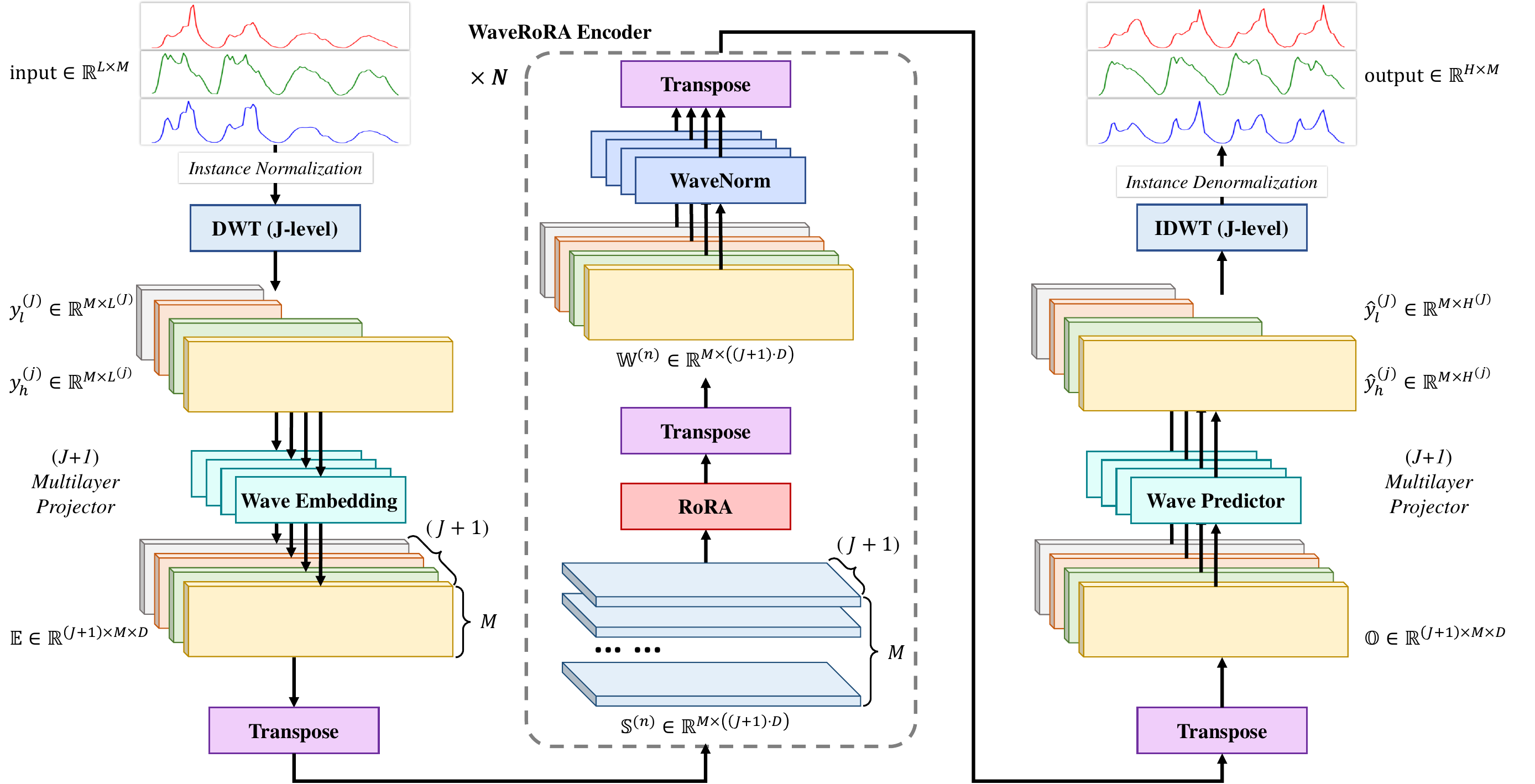}
    \caption{The architecture of WaveRoRA. The input MTS data is first stabilized by instance normalization and then transformed to multi-scale wavelet coefficients through $J$-level DWT. Each series of coefficients is passed through a corresponding wave embedding layer to generate uniformed wavelet-wise tokens. These embeddings are then transposed to series-wise tokens and fed into $N$-layer WaveRoRA Encoders to capture inter-series dependencies. Subsequently, the outputs are input to a set of wave predictors to obtain the predicted wavelet coefficients, which are then supplied to the IDWT to generate the final predictions.}
    \label{fig:arc}
\end{figure*}

\section{Methodology}
In this section, we overview the model and then detail the operations of each component in the model.
\subsection{Overview}
As shown in Fig \ref{fig:arc}, given the input MTS data $\mathbf{X}_{in}\in\mathbb{R}^{L\times M}$, we first apply instance normalization to stabilize the time series. We then apply $J$-level DWT decomposition and get $J$ high-pass coefficients $y_h^{(j)}\in\mathbb{R}^{M\times L^{(j)}},j\in\{1,2,...,J\}$ and 1 low-pass coefficients $y_l^{(J)}\in\mathbb{R}^{M\times L^{(J)}}$. Due to the downsampling during the multi-layer decomposition of DWT, the lengths of the wavelet coefficients of each layer are inconsistent. Therefore, we introduce a set of wave-embedding layers to map the multi-layer coefficients to wavelet-wise tokens with a uniformed size $D$, denoted as $\mathbb{E}\in\mathbb{R}^{(J+1)\times M\times D}$.
After the projection, we transpose $\mathbb{E}$ into series-wise tokens $\mathbb{S}\in\mathbb{R}^{M\times ((J+1)\cdot D)}$. We then use $N$-layer WaveRoRA Encoders to capture the inter-series dependencies in the wavelet domain.
% Specifically, $\mathbb{S}$ we transpose $\mathbb{S}$ again and apply wavelet-wise normalization. 
For the obtained features, we transpose them again to wavelet-wise tokens $\mathbb{O}\in\mathbb{R}^{(J+1)\times M\times D}$. Our objective is that each element in $\mathbb{O}$ approximates the high-pass and low-pass coefficients obtained from $J$-level wavelet decomposition of the ground truth. Therefore, we use a set of wave-predictors to map them to their corresponding lengths. Finally, we apply IDWT to get the predicted values.

\subsection{Instance Normalization}
Real-world series usually present non-stationarity\cite{liu2022non}, resulting in the distribution drift problem\cite{kim2021reversible}. Therefore, we adopt the instance normalization and denormalization technique that is used in Stationary Transformers\cite{liu2022non}. Specifically, we record the mean value $\mu_{in}$ and variance value $\sigma_{in}$ of the input series and use them to normalize the input series. When obtaining the output, we apply the same $\mu_{in}$ and $\sigma_{in}$ to denormalize the output series.

\subsection{Multi-scale Wave Embedding}
The input MTS data is first decomposed to $J$ high-pass coefficients $y_h^{(j)}\in\mathbb{R}^{M\times L^{(j)}}$ and 1 low-pass coefficients $y_l^{(J)}\in\mathbb{R}^{M\times L^{(J)}}$ with $J$-level DWT, where $j\in\{1,2,...,J\}$. Each coefficient series is passed through a corresponding wave-embedding layer defined in Eq \ref{eq::7}, where $\mathcal{C}=\{y_h^{(1)},y_h^{(2)}...,y_h^{(J)},y_l^{(J)}\}$ and $\mathrm{WaveEmbedding}$ is a linear projector. Specifically, $\mathrm{WaveEmbedding}_j$ contains a weight matrix $W_j$ and a bias matrix $b_j$. Note that $W_j\in\mathbb{R}^{D\times L^{(j)}}$ for $j\in\{1,2,...,J\}$ and $W_{J+1}\in\mathbb{R}^{D\times L^{(J)}}$, while $b_j\in\mathbb{R}^{D\times 1}$. The multi-scale wave-embedding layers unify the feature dimensions of wavelet coefficients obtained from the multi-layer DWT. What's more, the embedding process facilitates in capturing the temporal patterns within the time series.

\begin{equation}\label{eq::7}
\begin{aligned}
\mathbb{E}_j&=\mathrm{WaveEmbedding}_j(\mathcal{C}_j)\\
&=\mathcal{C}_jW_j^T+b_j^T
\end{aligned}
\end{equation}

\begin{figure*}[t]
\begin{equation}\label{eq::8}
\begin{aligned}
&\mathcal{M}_{\Theta,n}^{d}=\begin{pmatrix}
\cos{n\theta_1} & -\sin{n\theta_1} & 0 & 0 & \cdots & 0 & 0 \\
\sin{n\theta_1} & \cos{n\theta_1}  & 0 & 0 & \cdots & 0 & 0 \\
0 & 0 & \cos{n\theta_2} & -\sin{n\theta_2} & \cdots & 0 & 0 \\
0 & 0 & \sin{n\theta_2} & \cos{n\theta_2}  & \cdots & 0 & 0 \\
\vdots & \vdots & \vdots & \vdots & \ddots & \vdots & \vdots \\
0 & 0 & 0 & 0 & \cdots & \cos{n\theta_{\frac{d}{2}}} & -\sin{n\theta_{\frac{d}{2}}} \\
0 & 0 & 0 & 0 & \cdots & \sin{n\theta_{\frac{d}{2}}} & \cos{n\theta_{\frac{d}{2}}}
\end{pmatrix} 
\end{aligned}
\end{equation}
\end{figure*}

\subsection{WaveRoRA Encoder}
RoRA is proposed to capture inter-series dependencies in the wavelet domain. We first transpose $\mathbb{E}\in\mathbb{R}^{(J+1)\times M\times D}$ into series-wise tokens $\mathbb{S}\in\mathbb{R}^{M\times ((J+1)\cdot D)}$ where each token in $\mathbb{S}$ represents the concatenation of wavelet embeddings across all decomposition levels for a single variable. To simplify the representation, we use $D^\prime$ instead of $(J+1)\cdot D$ and $I\in\mathbb{R}^{M\times D^\prime}$ instead of $\mathbb{S}^{(n)}$ of each WaveRoRA Encoder.
\subsubsection{Rotary Route Attention}\label{sec:rora}
A set of linear projectors are applied to generate query, key and value matrices, represented as $Q\in\mathbb{R}^{M\times D^\prime}$, $K\in\mathbb{R}^{M\times D^\prime}$ and $V\in\mathbb{R}^{M\times D^\prime}$. Subsequently, a small number of routing tokens $R\in\mathbb{R}^{r\times D^\prime}$ is randomly initialized and the multi-head version of RoRA can be represented in Alg \ref{alg::8}, where $\odot$ refers to Hadamard product and the activation function $\sigma$ is set as $\mathrm{SiLU}$\cite{elfwing2018sigmoid}, generating $G\in\mathbb{R}^{M\times D^\prime}$ which serves as a gating function. The routing tokens $R$ is initially viewed as the query for $K$ and $V$, aggregating routing features from all keys and values. Afterward, $R$ and $V_R$ will serve as the key and value for $Q$ respectively, enabling the transfer of information from the routing features to each query token in $Q$. In our implementations, $R$ is expected to prune redundant relationships between low-correlation variables while preserving interaction information among high-correlation variables. In addition, we set $r\ll M$ as a fixed number to achieve linear complexity of $O(Mrd)$ relative to the variable number $M$.
% Subsequently, a rotary matrix $\mathcal{R}_\Theta^{D^\prime}$ will be applied to $Q$ and $K$ to inject relative positional information\cite{su2024roformer}. Specifically, $\mathcal{R}_\Theta^{D^\prime}$ consists of $M$ sub rotary matrices which can be defined in Eq \ref{eq::8}. We then randomly initialize a series of routing tokens $R\in\mathbb{R}^{r\times D^\prime}$ with a small number of $r$. The route attention can be represented in Eq \ref{eq::9}:

RoRA could be simplified to a form of generalized linear attention as shown in Eq \ref{eq::10}, which may lead to insufficient feature diversity\cite{han2023flatten}. Therefore, we try to solve this issue by injecting relative positional information through rotary mechanism. Typically, a rotary matrix $\mathcal{M}_{\Theta}^d$ defined in Eq \ref{eq::8} is multiplied with $Q$ and $K$ in SA or LA, where $\Theta$ represents a set of predefined rotational angles and $d$ is the dimension of each token. $\mathcal{M}_{\Theta}^d$ adaptively adjusts the angles based on each token's position, thereby incorporating the relative positional information $(m-n)\Theta$ for any indexes $m$, $n$ in $Q$ and $K$. However, such mechanism relies on the assumption that $Q$ and $K$ are directly multiplied to generate the attention scores. In our proposed RoRA, the routing tokens $R$ may disrupt the rotational angles applied to $Q$ and $K$. Assuming that we follow the typical use of RoPE as shown in Eq \ref{eq::9}, the relative positional information would not be correctly computed. Specifically, the hypothetical rotary attention can be expressed as Eq \ref{eq::10}, where $\Psi$ could be a certain kernel function. Assuming that $\Psi$ satisfies linear properties, we can then get an approximate solution $\Psi(Q\mathcal{M}^{D^\prime}_{\Theta} R^TR(\mathcal{M}^{D^\prime}_{\Theta})^TK^T)$. Furthermore, it can easily be demonstrated that for any position indexes $m$ and $n$, the relative positional embeddings with the rotational angle $(m-n)\Theta$ could only be correctly obtained when $R^TR$ equals to an  identity matrix. However, the routing tokens are randomly initialized and $\Psi$ could also be non-linear functions like Softmax. Therefore, applying rotary matrix to $Q$ and $K$ may not properly model the relative positional information.

\begin{equation}\label{eq::9}
\begin{aligned}
V_R^\prime&=\mathrm{Attention}(R,K\mathcal{M}^{D^\prime}_{\Theta},V)\\
V_O^\prime&=\mathrm{Attention}(Q\mathcal{M}^{D^\prime}_{\Theta},R,V_R)
\end{aligned}
\end{equation}

\begin{equation}\label{eq::10}
\begin{aligned}
V_O^\prime&=\Psi(Q\mathcal{M}^{D^\prime}_{\Theta} R^T)\Psi(R(K\mathcal{M}^{D^\prime}_{\Theta})^T)V\\
&\approx \Psi\big(Q\mathcal{M}^{D^\prime}_{\Theta} R^TR(\mathcal{M}^{D^\prime}_{\Theta})^TK^T\big)V
\end{aligned}
\end{equation}

\begin{figure}[t]
    \centering
    \includegraphics[width=1.0\linewidth]{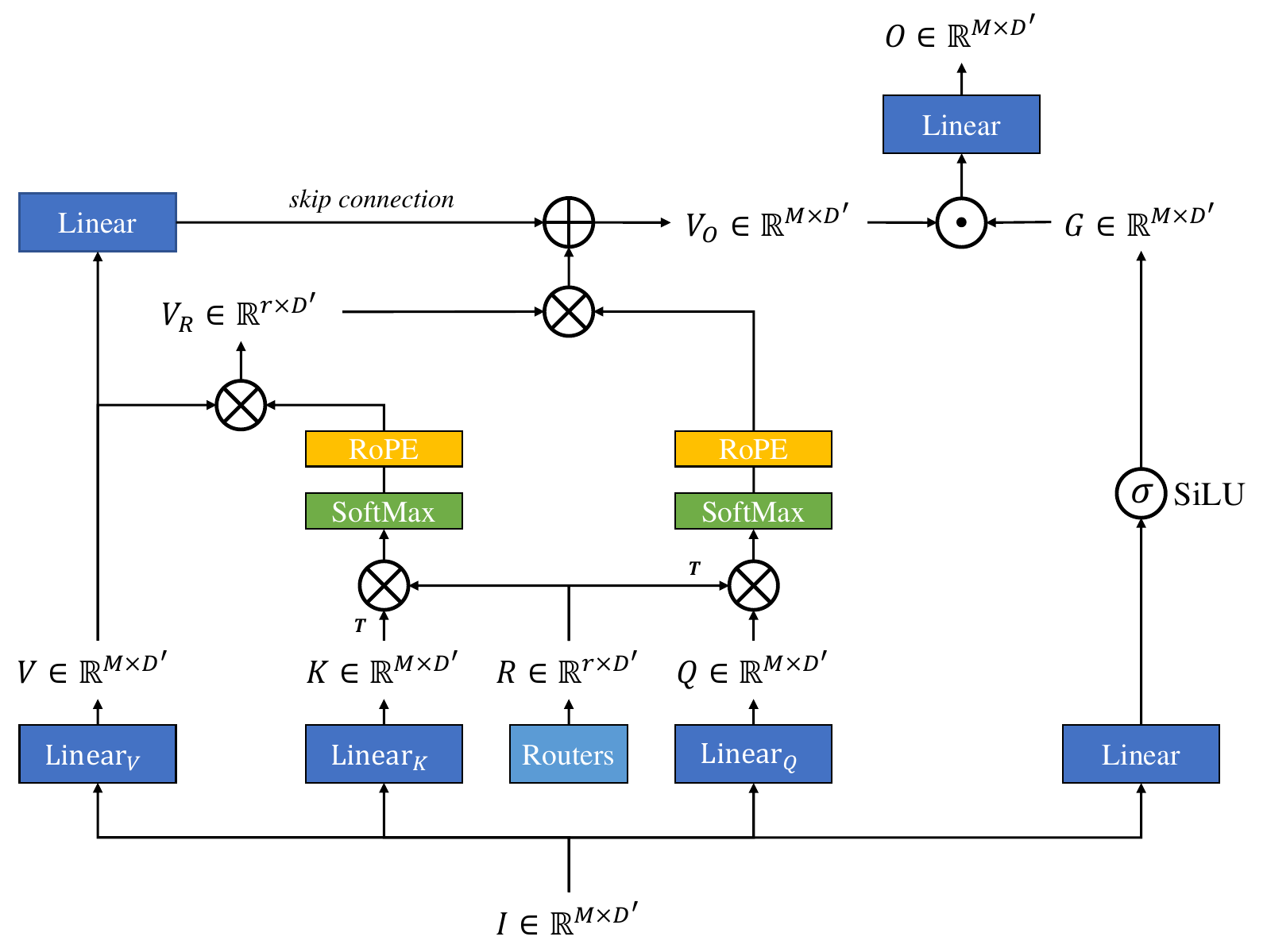}
    \caption{The architecture of RoRA.}
    \label{fig:RoRA}
\end{figure}

\begin{algorithm}[t]
\caption{The implementation of multi-head RoRA}\label{alg::8}
\renewcommand{\algorithmicrequire}{\textbf{Input:}}
\renewcommand{\algorithmicensure}{\textbf{Output:}}
\begin{algorithmic}[1]
\REQUIRE Input sequence tokens $I\in\mathbb{R}^{M\times D^\prime}$
\ENSURE Output representations $O\in\mathbb{R}^{M\times D^\prime}$
\setstretch{1.1}
\STATE $G\in\mathbb{R}^{N\times D^\prime}\leftarrow\sigma\big(\mathrm{Linear}(I)\big)$
\STATE $R\in\mathbb{R}^{r\times D^\prime}\leftarrow$ Randomly initialized 
\FOR{$h=1$ to $\mathrm{H}$}
\STATE $Q^h\in\mathbb{R}^{M\times (D^\prime/H)}\leftarrow\mathrm{Linear}(I)$
\STATE $K^h\in\mathbb{R}^{M\times (D^\prime/H)}\leftarrow\mathrm{Linear}(I)$
\STATE $V^h\in\mathbb{R}^{M\times (D^\prime/H)}\leftarrow\mathrm{Linear}(I)$
\STATE $R^h\in\mathbb{R}^{r\times (D^\prime/H)}\leftarrow\mathrm{Linear}(R)$
\STATE $A_{R-K}^h\in\mathbb{R}^{r\times M}\leftarrow\big(\Psi(RK^T)\mathcal{M}^{r}_{\Theta}\big)^T$
\STATE $V_R^h\in\mathbb{R}^{r\times (D^\prime/H)}\leftarrow A_{R-K}^hV^h$
\STATE $A_{Q-R}^h\in\mathbb{R}^{M\times r}\leftarrow \Psi(QR^T)\mathcal{M}^{r}_{\Theta}$
\STATE $V_O^h\in\mathbb{R}^{M\times (D^\prime/H)}\leftarrow A_{Q-R}^hV_R^h+\mathrm{Linear}(V^h)$
\ENDFOR
\STATE $O\in\mathbb{R}^{M\times D^\prime}\leftarrow\mathrm{Linear}\big(\mathrm{Concat}(V_O^1,V_O^2,...,V_O^H)\odot G\big)$
\STATE \textbf{return} $O$
\end{algorithmic}
\end{algorithm}

Based on the conclusions, we apply $\mathcal{M}^{r}_{\Theta}$ to the attention scores of $Q$-$R$ and $R$-$K$ pairs, as shown in Alg \ref{alg::8}. The basic formula of RoRA can then be written as Eq \ref{eq::11}, where $\Psi$ here is the Softmax function. In this case, the rotary matrix properly reflect the relative positional differences between elements. Although the corresponding dimensionality is reduced to $r$, RoRA could take advantages of the nonlinearity of Softmax and the mapping mechanism of $R$. In addition, we introduce a lienar projector as the skip-connection module to further enhance the feature diversity of $Q$ and $K$. 

\begin{equation}\label{eq::11}
\begin{aligned}
V_O=&\Psi(QR^T)\mathcal{M}^{r}_{\Theta}(\mathcal{M}^{r}_{\Theta})^T\Psi(RK^T)^TV+\mathrm{Linear}(V)
\end{aligned}
\end{equation}

\subsubsection{Wavelet-wise Normalization}
Considering that the wavelet coefficients of different decomposition layers correspond to different frequency intervals, we transpose the output of RoRA to wavelet-wise tokens again and apply wavelet-wise normalization. The process is defined in Eq \ref{eq::12}, where $\mathrm{RoRA}$ corresponds to Alg \ref{alg::8} , $\mathbb{S}^{(n)}\in\mathbb{R}^{M\times ((J+1)\cdot D)}$ is the series-wise tokens, $\mathbb{W}^{(n)}\in\mathbb{R}^{(J+1)\times M\times D}$ is the wavelet-wise tokens and $n$ represents the number of WaveRoRA Encoder layer.

\begin{equation}\label{eq::12}
\begin{aligned}
\mathbb{W}^{(n)}&=\mathrm{Transpose}(\mathrm{RoRA}(\mathbb{S}^{(n)}))\\
\mathbb{S}^{(n+1)}&=\mathrm{Transpose}(\mathrm{WaveNorm}(\mathbb{W}^{(n)}))
\end{aligned}
\end{equation}

\subsection{Multi-scale Wavelet Predictor}
In order to reconstruct the predicted series with length $H$, the length of high-pass and low-pass coefficients of each decomposition layers input to the IDWT should adhere to the same rules as DWT. Specifically, for $\hat{y}_h^{(j)}\in\mathbb{R}^{M\times H^{(j)}}$ and $\hat{y}_l^{(J)}\in\mathbb{R}^{M\times H^{(J)}}$, we set $H^{(j)}=\lfloor\frac{H^{(j-1)}+S-1}{2}\rfloor$ and $H^{(0)}=H$. To map the embeddings to the specific $H^{(j)}$, we transpose the output of the last RoRA layer to wavelet embeddings $\mathbb{O}\in\mathbb{R}^{(J+1)\times M\times D}$ and pass them through a corresponding MLP individually, defined as Eq \ref{eq::13}:
\begin{equation}\label{eq::13}
\begin{aligned}
\widehat{\mathcal{C}}_j&=\mathrm{WavePredictor}_j(\mathbb{O}_j)\\
&=(\mathrm{GELU}(\mathbb{O}_j))\hat{W}_j^T+\hat{b}_j^T
\end{aligned}
\end{equation}
where $\widehat{\mathcal{C}}=\{\hat{y_h}^{(1)},\hat{y_h}^{(2)}...,\hat{y_h}^{(J)},\hat{y_l}^{(J)}\}$ contains the predicted high-pass and low-pass coefficients, $W_j\in\mathbb{R}^{H^{(j)}\times D}$ and $b_j\in\mathbb{R}^{H^{(j)}\times 1}$ for $j\in\{1,2,...,J\}$, while $W_{J+1}\in\mathbb{R}^{H^{(J)}\times D}$ and $b_{J+1}\in\mathbb{R}^{H^{(J)}\times 1}$. We set $\widehat{\mathcal{C}}$ as the input of IDWT to get the final outputs $X_{out}\in\mathbb{R}^{H\times M}$.

\section{Experiments}
\subsection{Datasets}
We evaluate our model on 8 real-world datasets: Weather, Electricity, Traffic, Solar and 4 ETT datasets (ETTh1, ETTh2, ETTm1, ETTm2)\cite{liu2023itransformer}. These datasets are extensively utilized, covering multiple fields including weather, energy management and transportation. The statistics of the datasets are shown in Table \ref{tab:data}.

\begin{table*}[t]
    \centering
    \caption{Statistics of all datasets.}
    \label{tab:data}
    \scalebox{0.95}{
        \begin{tabular}{ccccccccc}
        \toprule 
        Dataset & Weather & Electricity & Traffic & ETTh1 & ETTh2 & ETTm1 & ETTm2 & Solar \\
        Variables & 21  & 321 & 862 & 7 & 7 & 7 & 7 & 137 \\
        Frequency & 10 min & 1 hour & 1 hour & 1 hour & 1 hour & 15 min & 15 min & 10 min \\
        \bottomrule 
        \end{tabular}
    }
\end{table*}

\begin{table*}[t]
\centering
\caption{Experimental results of MTS time series forecasting task on 8 real-world datasets. The best results are in \textbf{bold}.}
\label{tab:result total}
\renewcommand{\arraystretch}{1.2}
\resizebox{0.8\linewidth}{!}{
    \begin{tabular}{cc|c|cc|cc|cc|cc|cc|cc|ccc}
        \cline{2-17}
        &\multicolumn{2}{c|}{Models}& \multicolumn{2}{c|}{WaveRoRA}& \multicolumn{2}{c|}{iTransformer}& \multicolumn{2}{c|}{PatchTST}& \multicolumn{2}{c|}{Crossformer}& \multicolumn{2}{c|}{Autoformer}& \multicolumn{2}{c|}{Dlinear} & \multicolumn{2}{c}{TimesNet} \\
        \cline{2-17}
        &\multicolumn{2}{c|}{Metric}&MSE&MAE&MSE&MAE&MSE&MAE&MSE&MAE&MSE&MAE&MSE&MAE&MSE&MAE\\
        \cline{2-17}
        &\multirow{4}*{\rotatebox{90}{Weather}}& 96    
        & 0.159 & \textbf{0.204} & 0.175 & 0.216 & 0.172 & 0.214 & \textbf{0.158} & 0.230 & 0.249 & 0.329 & 0.198 & 0.260 & 0.172 & 0.220 \\
        &\multicolumn{1}{c|}{}& 192   
        & 0.208 & \textbf{0.250} & 0.228 & 0.261 & 0.218 & 0.256 & \textbf{0.206} & 0.277 & 0.325 & 0.370 & 0.240 & 0.301 & 0.219 & 0.261 \\
        &\multicolumn{1}{c|}{}& 336   
        & \textbf{0.263} & \textbf{0.292} & 0.282 & 0.300 & 0.275 & 0.297 & 0.272 & 0.335 & 0.351 & 0.391 & 0.287 & 0.340 & 0.280 & 0.306 \\
        &\multicolumn{1}{c|}{}& 720   
        & \textbf{0.345} & \textbf{0.346} & 0.360 & 0.350 & 0.352 & 0.347 & 0.398 & 0.418 & 0.415 & 0.426 & 0.353 & 0.395 & 0.365 & 0.359 \\
        \cline{2-17}&\multirow{4}*{\rotatebox{90}{Traffic}}& 96    
        & \textbf{0.364} & \textbf{0.250} & 0.392 & 0.263 & 0.435 & 0.274 & 0.522 & 0.290 & 0.597 & 0.371 & 0.652 & 0.386 & 0.593 & 0.321 \\
        &\multicolumn{1}{c|}{} & 192   
        & \textbf{0.376} & \textbf{0.258} & 0.414 & 0.272 & 0.444 & 0.280 & 0.530 & 0.293 & 0.607 & 0.382 & 0.601 & 0.373 & 0.617 & 0.336 \\
        &\multicolumn{1}{c|}{}& 336   
        & \textbf{0.389} & \textbf{0.270} & 0.430 & 0.283 & 0.455 & 0.285 & 0.558 & 0.305 & 0.623 & 0.387 & 0.605 & 0.373 & 0.629 & 0.336 \\
        &\multicolumn{1}{c|}{}& 720   
        & \textbf{0.425} & \textbf{0.285} & 0.452 & 0.297 & 0.488 & 0.304 & 0.589 & 0.328 & 0.639 & 0.395 & 0.649 & 0.399 & 0.640 & 0.350 \\
        \cline{2-17}
        &\multirow{4}*{\rotatebox{90}{Electricity}}& 96    
        & \textbf{0.136} & \textbf{0.231} & 0.153 & 0.245 & 0.167 & 0.253 & 0.219 & 0.314 & 0.196 & 0.313 & 0.195 & 0.278 & 0.168 & 0.272 \\
        &\multicolumn{1}{c|}{}& 192   
        & \textbf{0.153} & \textbf{0.247} & 0.167 & 0.256 & 0.175 & 0.261 & 0.231 & 0.322 & 0.211 & 0.324 & 0.195 & 0.281 & 0.184 & 0.289 \\
        &\multicolumn{1}{c|}{}& 336   
        & \textbf{0.169} & \textbf{0.266} & 0.181 & 0.272 & 0.190 & 0.277 & 0.246 & 0.337 & 0.214 & 0.327 & 0.207 & 0.296 & 0.198 & 0.300 \\
        &\multicolumn{1}{c|}{}& 720   
        & \textbf{0.196} & \textbf{0.291} & 0.219 & 0.305 & 0.236 & 0.314 & 0.280 & 0.363 & 0.236 & 0.342 & 0.242 & 0.328 & 0.220 & 0.320 \\
        \cline{2-17}
        &\multirow{4}*{\rotatebox{90}{ETTh1}}& 96    
        & 0.381 & 0.402 & 0.394 & 0.410 & \textbf{0.380} & \textbf{0.399} & 0.423 & 0.448 & 0.435 & 0.446 & 0.391 & 0.403 & 0.384 & 0.402 \\
        &\multicolumn{1}{c|}{}& 192   
        & \textbf{0.425} & 0.429 & 0.443 & 0.437 & \textbf{0.425} & \textbf{0.427} & 0.471 & 0.474 & 0.456 & 0.457 & 0.444 & 0.438 & 0.436 & 0.429 \\
        &\multicolumn{1}{c|}{}& 336   
        & 0.466 & 0.447 & 0.485 & 0.460 & \textbf{0.464} & \textbf{0.445} & 0.570 & 0.546 & 0.486 & 0.487 & 0.480 & 0.463 & 0.491 & 0.469 \\
        &\multicolumn{1}{c|}{}& 720   
        & \textbf{0.458} & \textbf{0.464} & 0.504 & 0.490 & 0.485 & 0.469 & 0.653 & 0.621 & 0.515 & 0.517 & 0.513 & 0.508 & 0.521 & 0.500 \\
        \cline{2-17}
        &\multirow{4}*{\rotatebox{90}{ETTh2}}& 96    
        & \textbf{0.281} & \textbf{0.337} & 0.303 & 0.353 & 0.293 & 0.342 & 0.745 & 0.584 & 0.332 & 0.368 & 0.345 & 0.397 & 0.340 & 0.374 \\
        &\multicolumn{1}{c|}{}& 192   
        & \textbf{0.356} & \textbf{0.387} & 0.385 & 0.401 & 0.383 & 0.396 & 0.877 & 0.656 & 0.426 & 0.434 & 0.473 & 0.472 & 0.402 & 0.414 \\
        &\multicolumn{1}{c|}{}& 336   
        & \textbf{0.394} & \textbf{0.416} & 0.425 & 0.433 & 0.423 & 0.431 & 1.043 & 0.731 & 0.477 & 0.479 & 0.584 & 0.535 & 0.452 & 0.452 \\
        &\multicolumn{1}{c|}{}& 720   
        & \textbf{0.403} & \textbf{0.433} & 0.434 & 0.449 & 0.424 & 0.442 & 1.104 & 0.763 & 0.453 & 0.490 & 0.795 & 0.642 & 0.462 & 0.468 \\
        \cline{2-17}
        &\multirow{4}*{\rotatebox{90}{ETTm1}}& 96    
        & \textbf{0.320} & \textbf{0.357} & 0.339 & 0.374 & 0.326 & 0.363 & 0.404 & 0.426 & 0.510 & 0.492 & 0.344 & 0.372 & 0.338 & 0.375 \\
        &\multicolumn{1}{c|}{}& 192   
        & 0.365 & \textbf{0.383} & 0.380 & 0.394 & \textbf{0.362} & 0.385 & 0.450 & 0.451 & 0.514 & 0.495 & 0.383 & 0.393 & 0.374 & 0.387 \\
        &\multicolumn{1}{c|}{}& 336   
        & 0.396 & \textbf{0.404} & 0.417 & 0.418 & \textbf{0.393} & \textbf{0.409} & 0.532 & 0.515 & 0.521 & 0.497 & 0.414 & 0.416 & 0.410 & 0.411 \\
        &\multicolumn{1}{c|}{}& 720   
        & 0.463 & 0.442 & 0.478 & 0.452 & \textbf{0.451} & \textbf{0.439} & 0.666 & 0.589 & 0.537 & 0.513 & 0.474 & 0.452 & 0.478 & 0.450 \\
        \cline{2-17}
        &\multirow{4}*{\rotatebox{90}{ETTm2}} & 96    
        & \textbf{0.175} & \textbf{0.259} & 0.189 & 0.274 & 0.177 & 0.261 & 0.287 & 0.366 & 0.205 & 0.293 & 0.190 & 0.287 & 0.187 & 0.267 \\
        &\multicolumn{1}{c|}{}& 192   
        & \textbf{0.241} & \textbf{0.302} & 0.254 & 0.313 & 0.247 & 0.309 & 0.414 & 0.492 & 0.278 & 0.336 & 0.274 & 0.349 & 0.249 & 0.309 \\
        &\multicolumn{1}{c|}{}& 336   
        & \textbf{0.304} & \textbf{0.342} & 0.315 & 0.352 & 0.312 & 0.351 & 0.597 & 0.542 & 0.343 & 0.379 & 0.376 & 0.421 & 0.321 & 0.351 \\
        &\multicolumn{1}{c|}{}& 720   
        & \textbf{0.405} & \textbf{0.398} & 0.413 & 0.405 & 0.424 & 0.416 & 1.073 & 1.042 & 0.414 & 0.419 & 0.545 & 0.519 & 0.408 & 0.403 \\
        \cline{2-17}
        &\multirow{4}*{\rotatebox{90}{Solar}}& 96    
        & \textbf{0.197} & \textbf{0.222} & 0.207 & 0.241 & 0.202 & 0.240 & 0.310 & 0.331 & 0.864 & 0.707 & 0.284 & 0.372 & 0.250 & 0.292 \\
        &\multicolumn{1}{c|}{}& 192   
        & \textbf{0.232} & \textbf{0.254} & 0.240 & 0.264 & 0.240 & 0.265 & 0.734 & 0.725 & 0.844 & 0.695 & 0.320 & 0.396 & 0.296 & 0.318 \\
        &\multicolumn{1}{c|}{}& 336   
        & \textbf{0.250} & \textbf{0.275} & 0.251 & 0.274 & 0.249 & 0.276 & 0.750 & 0.735 & 0.751 & 0.700 & 0.354 & 0.421 & 0.319 & 0.330 \\
        &\multicolumn{1}{c|}{}& 720   
        & \textbf{0.247} & \textbf{0.274} & 0.253 & 0.277 & 0.248 & 0.277 & 0.769 & 0.765 & 0.867 & 0.710 & 0.354 & 0.415 & 0.338 & 0.337 \\
        \cline{2-17}
    \end{tabular}
}
\end{table*}

\subsection{Baselines}
We select 6 baseline methods for the comparative experiments, including 4 Transformer-based models: iTransformer\cite{liu2023itransformer}, PatchTST\cite{nie2023patch}, Crossformer\cite{zhang2023crossformer}, Auto-former\cite{wu2021autoformer}; 1 CNN-based method: TimesNet\cite{wu2023timesnet} and 1 MLP-based method: DLinear\cite{zeng2023transformers}. 

\subsection{Experiment Settings}\label{sec:setting}
We set a uniform input length $L=96$ for all models and conduct experiments with $H=\{96,192,336,720\}$. By default, WaveRoRA employs a 4-level DWT using the Symlet3\cite{daubechies1992ten} wavelet basis. We map the wavelet coefficients to embeddings with $D=64$. For datasets with less variables, such as ETT datasets, we set RoRA layer number $N$ to 2; while for those with more variables like Weather, Electricity and Solar, we set $N=3$; For Traffic that has large number of variables, we set $N=4$. For RoRA, we set the token number of $R$ to $r=\min(10,\frac{\log{M}+\sqrt{M}}{2})$. The attention heads number is set to 8 by default.

We replicate and run PatchTST, iTransformer and DLinear using the publicly available optimal parameters. For other baseline models, we adhere to their established implementations. We record the MSE and MAE to measure the performance of different models. These metrics are defined in Eq \ref{eq::mse} and Eq \ref{eq::mae}, where $Y$ and $\hat{Y}$ represent ground truth and predicted values respectively. All experiments are conducted on one NVIDIA A800 GPU with 40GB memory to exclude the influence of platforms.

\begin{align}
    \mathrm{MSE}(Y,\hat{Y}) &= \frac{1}{|Y|}\sum_{i=1}^{|Y|}|Y_i-\hat{Y}_i|^2 \label{eq::mse} \\
    \mathrm{MAE}(Y,\hat{Y}) &= \frac{1}{|Y|}\sum_{i=1}^{|Y|}|Y_i-\hat{Y}_i| \label{eq::mae}
\end{align}

\begin{table*}[t]
\centering
\caption{Ablation results of MTS forecasting task on Weather, Traffic and Electricity. Each MSE and MAE score is the average value of $H=\{96,192,336,720\}$. The best results are in \textbf{bold}.}
\label{tab:ablation}
\renewcommand{\arraystretch}{1.2}
\resizebox{0.64\linewidth}{!}{
\setlength{\tabcolsep}{2.5pt}
    \begin{tabular}{cc|cc|cc|cc|cc|cc|cc|cc}
        \toprule
        \multicolumn{2}{c|}{Models}& \multicolumn{2}{c|}{w/ SA}& \multicolumn{2}{c|}{w/ LA}& \multicolumn{2}{c|}{w/o Ro}& \multicolumn{2}{c}{w/o Gate} & \multicolumn{2}{c}{w/o skip} & \multicolumn{2}{c}{WaveRoRA} & \multicolumn{2}{c}{iTransformer}\\
        \cmidrule(lr){1-2}\cmidrule(lr){3-4}\cmidrule(lr){5-6}\cmidrule(lr){7-8}\cmidrule(lr){9-10}\cmidrule(lr){11-12}\cmidrule(lr){13-14}\cmidrule(lr){15-16}
        \multicolumn{2}{c|}{Metric}&MSE&MAE&MSE&MAE&MSE&MAE&MSE&MAE&MSE&MAE&MSE&MAE&MSE&MAE\\
        \midrule
        \multicolumn{2}{c|}{Traffic} & 0.400 & 0.275 & 0.426 & 0.280 & 0.396 & 0.271 & 0.405 & 0.276 & 0.393 & 0.270 & \textbf{0.389} & \textbf{0.269} & 0.422 & 0.279 \\
        \midrule
        \multicolumn{2}{c|}{Electricity} & 0.168 & 0.265 & 0.174 & 0.267 & 0.171 & 0.266 & 0.174 & 0.269 & 0.168 & 0.264 & \textbf{0.165} & \textbf{0.261} & 0.180 & 0.270 \\
        \midrule
        \multicolumn{2}{c|}{ETTh1} & 0.454 & 0.448 & 0.452 & 0.446 & 0.444 & 0.446 & 0.450 & 0.445 & 0.455 & 0.447 & \textbf{0.433} & \textbf{0.436} & 0.457  & 0.449 \\
        \midrule
        \multicolumn{2}{c|}{ETTh2} & 0.367 & 0.402 & 0.364 & 0.398 & 0.365 & 0.396 & 0.369 & 0.398 & 0.371 & 0.400 & \textbf{0.359} & \textbf{0.393} & 0.387 & 0.409 \\
        \bottomrule
    \end{tabular}
}
\end{table*}

\subsection{Main Results}
Table \ref{tab:result total} presents all results of MTS forecasting. Compared with the current SOTA Transformer-based model, namely iTransformer, WaveRoRA reduces MSE by 5.91\% and MAE by 3.50\%. Among all the datasets, WaveRoRA achieves the most improvement on Traffic and Electricity, which contain large amount of variables and exhibit more pronounced periodic characteristics. In particular, the reductions of MSE and MAE come to 7.94\% and 4.66\% on Traffic and 9.17\% and 3.99\% on Electricity respectively. Such degree of improvement can be attributed to two aspects: (a) The wavelet transform can accurately model varying periodic patterns over time; (b) The proposed RoRA prunes redundant inter-series dependencies and retain key information from variables with genuine interactions.

For ETT datasets with less relevancy and non-significant periodicity, capturing inter-series dependencies may negatively impact the predicting accuracy. Models adopting the channel-independent strategy usually achieve better results, such as PatchTST and TimeNet. However, WaveRoRA also shows comparable performance on ETTh1 and ETTm1, while achieving more accurate predictions on ETTh2 and ETTm2. Compared to PatchTST, the average improvements of MSE and MAE are 2.36\% and 1.39\%. This indicates that wavelet transform can effectively preserve time-frequency characteristics while retaining the benefits of time-domain modeling. In addition, we believe that high-quality relative positional information facilitates enhancing interactions between highly correlated variables while weakening those between less correlated ones.

\subsection{Ablation Study}
To justify the effectiveness of our designs, we conduct additional experiments on Traffic, Electricity, ETTh1 and ETTh2 and accord the average MSE and MAE. The results are shown in Table \ref{tab:ablation}.
\subsubsection{Wavelet Learning Framework}
We replace RoRA with other existing backbones to capture inter-series dependencies, which corresponds to a) w/ SA which uses Softmax attention. Since the wavelet domain embeddings are also generated through an MLP structure, w/ SA can be viewed as replacing the temporal features of iTransformer with wavelet domain features. Compared with iTransformer, w/ SA reduces the MSE by 4.40\% and MAE by 1.36\% on average, which indicates that the wavelet learning framework could provide better predicting results.

\begin{figure}[t]
    \centering
    \begin{minipage}[b]{0.49\linewidth}
        \centering
        \subfigure[Hyper-parameter Analysis of \newline$J$ and dropout values on ETTh1]{
        \includegraphics[width=\linewidth]{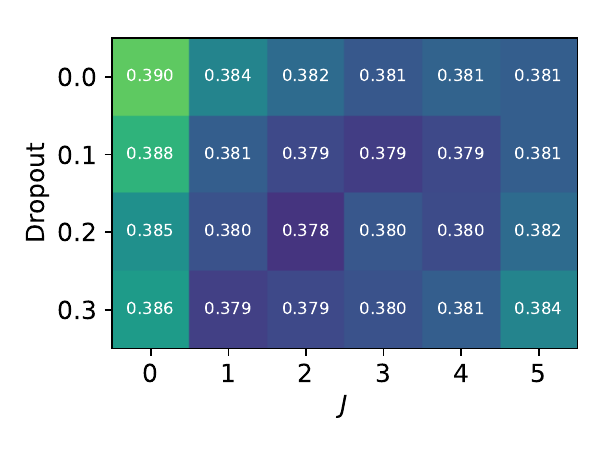}
        }
    \end{minipage}
    \hfill
    \begin{minipage}[b]{0.49\linewidth}
        \centering
        \subfigure[Hyper-parameter Analysis of \newline$J$ and dropout values on ETTh2]{
        \includegraphics[width=\linewidth]{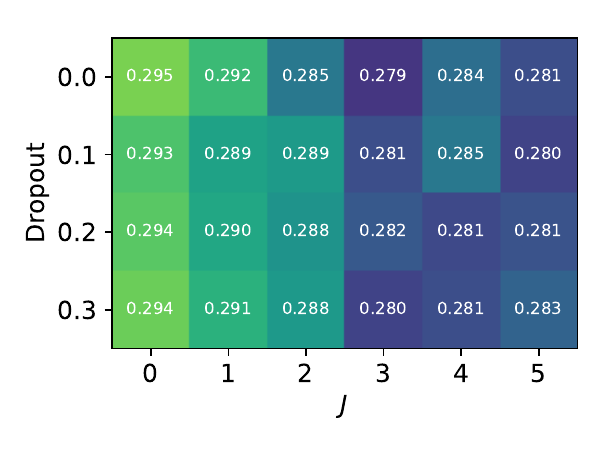}
        }
    \end{minipage}
    \caption{The MSE results with prediction length $H=96$ on ETTh1 and ETTh2. Varying settings of DWT decomposition level $J$ and dropout values are applied. The model's prediction accuracy improves and stabilizes with a larger $J$, while different dropout values have a minimal effect on performance.}
    \label{fig:JD}
\end{figure}

\begin{figure}[!htbp]
    \centering
    \subfigure[Hyper-parameter Analysis of $r$ on ETTh1]{\includegraphics[width=0.47\linewidth]{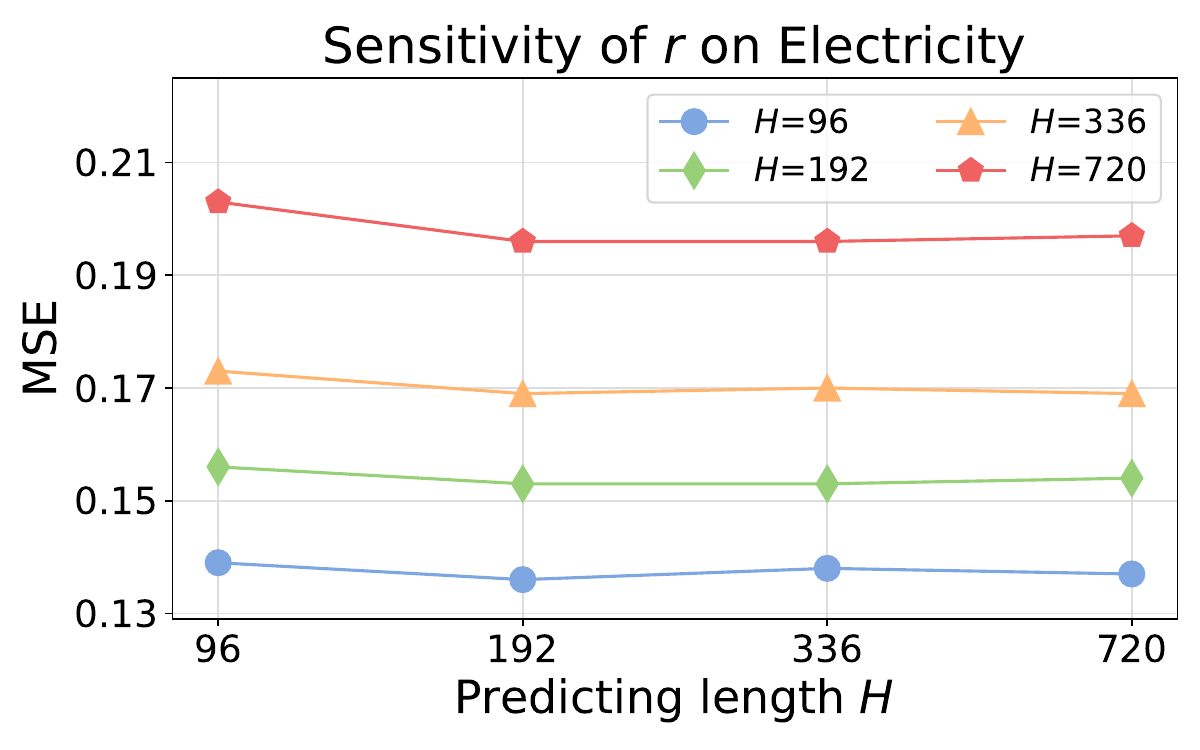}}
    \hfill
    \subfigure[Hyper-parameter Analysis of $r$ on ETTh2]{\includegraphics[width=0.47\linewidth]{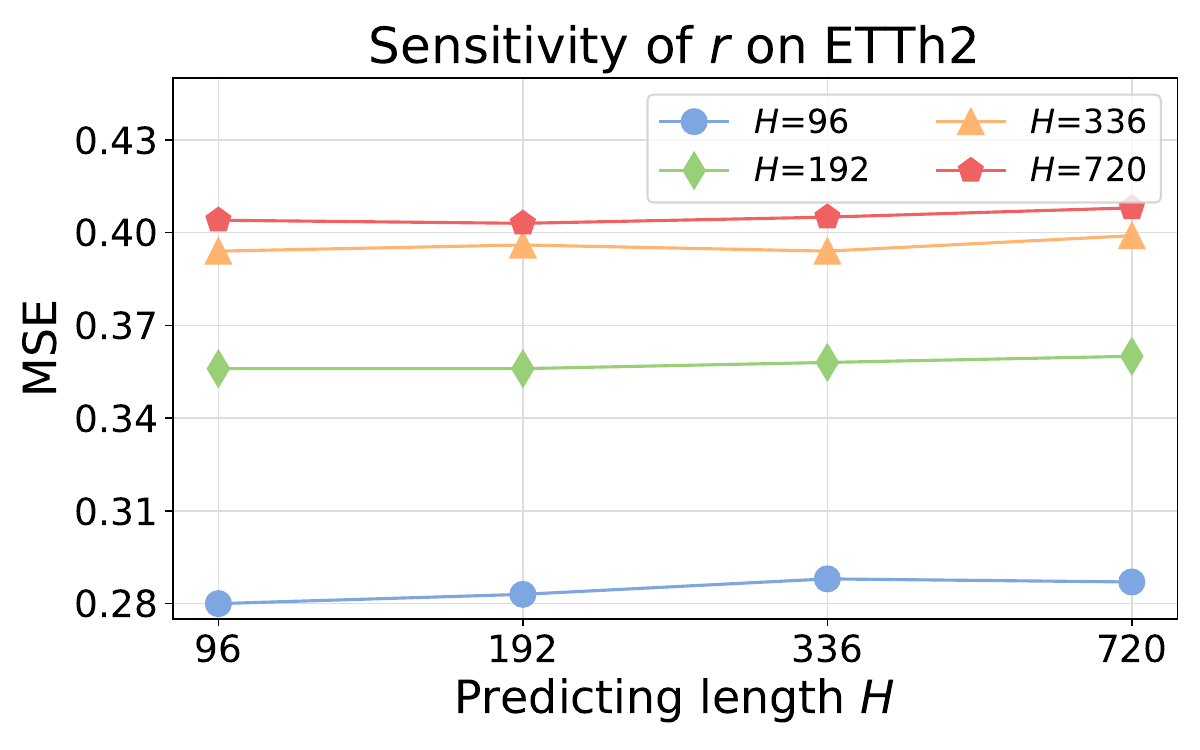}}
    \caption{The MSE results with prediction length $H=\{96,192,336,720\}$ on Electricity and ETTh2.}
    \label{fig:r}
\end{figure}

\subsubsection{RoRA structure}
To evaluate the effectiveness of the overall structure of RoRA, we further conduct b) w/ LA which uses linear attention to replace RoRA. SA and LA are typical representatives of the attention family, with various attention mechanisms essentially being derivatives of the two. It can be summarized from Table \ref{tab:ablation} that w/ SA and w/ LA achieve suboptimal results compared to WaveRoRA, indicating that RoRA has superior  representational capacity.

We then conduct c) w/o Ro which removes the rotary positional embeddings and d) w/o skip which removes the skip connection design to evaluate the components of RoRA. After these modules are removed, the predictive performance of the model decreases to varying degrees. The performance decline is most significant in w/o Gate on the Traffic and Electricity datasets, where the MSE and MAE increase by 5.03\% and 3.81\% on average. However, for ETTh1 and ETTh2, removing the skip connection layer causes more deterioration in the prediction accuracy, where the MSE and MAE increase by an average of 4.25\% and 1.95\%. For datasets with a larger number of variables, the feature diversity carried by series-wise tokens is generally guaranteed, making it beneficial to select high-quality information for more accurate predictions. In contrast, for datasets with fewer variables, enhancing the feature diversity often leads to improved performance.

% While the routing tokens contribute to prune redundant information, the gating unit further filter out the interaction between weakly correlated variables.

\begin{figure*}[t]
    \centering
    \subfigure[Wavelet functions of Symlet, Coiflet and Haar.]{\includegraphics[width=0.85\linewidth]{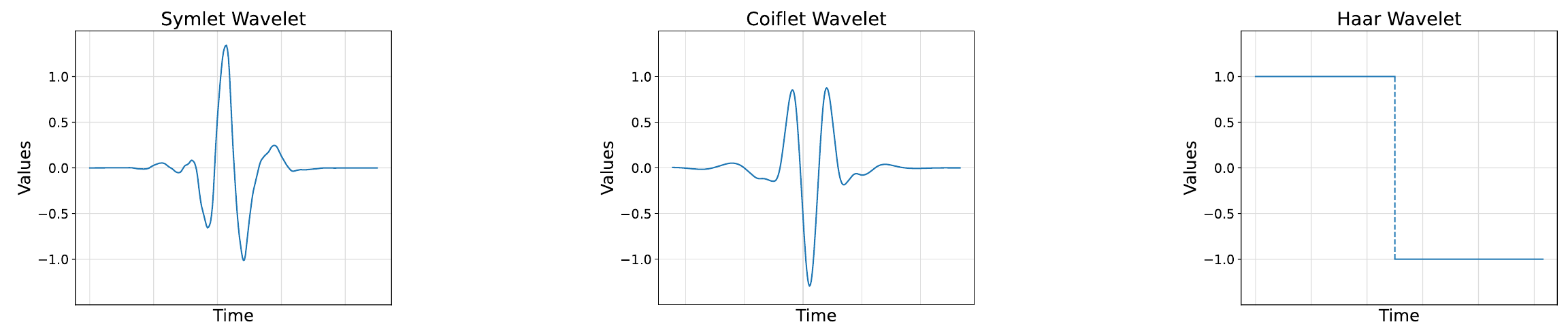}}
    \hfill
    \subfigure[The average MSE on different datasets.]{\includegraphics[width=0.85\linewidth]{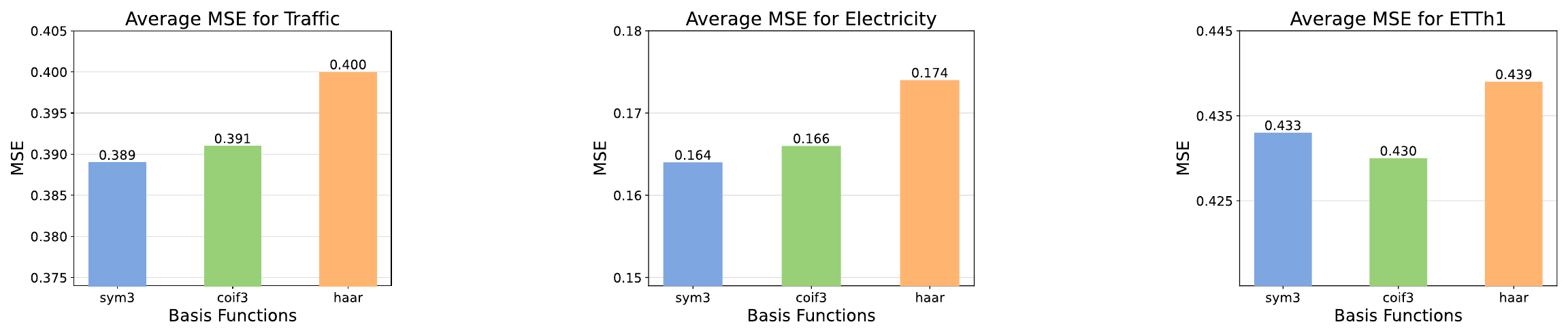}}
    \caption{The figures of (a) specific shape of Symlet, Coiflet and Haar wavelets and (b) average MSE results with prediction length $H=\{96,192,336,720\}$ on Traffic, Electricity and ETTh2.}
    \label{fig:basis}
\end{figure*}

\subsection{Hyper-parameter Sensitivity Analysis}
\subsubsection{DWT decomposition levels \& Dropout}
DWT hierarchically decomposes a time series into low-pass and high-pass components, corresponding to coarse trends and finer periodical details at different scales. With the increase of decomposition level $J$, coarser periodic information is separated and the low-pass component will exhibit more significant trend characteristics. To obtain more comprehensive results, we change the value of dropout layers at the same time. Dropout layers improve the generalization ability of the model by randomly masking part of the input. A proper dropout value is conducive to avoiding overfitting. The dropout layer mainly works in each \textbf{WaveNorm} layer. We set $J=\{0,1,2,3,4,5\}$ and $dropout=\{0.0,0.1,0.2,0.3\}$. Note that when $J=0$, the model does not make any DWT decomposition and RoRA is applied to time-domain features. We conduct the experiments on ETTh1 and ETTh2 and the final results are shown in Fig \ref{fig:JD}. Generally, different dropout values do not cause substantial changes of model performance. In contrast, whether to operate DWT to use wavelet-domain features could lead to significant variations in the results. Specifically, when $J=0$, i.e. only time-domain features are used, the model provides suboptimal results. With the increase of $J$, the model's prediction accuracy improves and stabilizes. This indicates that key periodic components is separated and the remaining trend components no longer contain extra periodic information.

\subsubsection{Number of routing tokens}
Typically, the routing tokens are kept with $r\ll M$. A larger $r$ could lead to more comprehensive capturing of element-wise correlation but may also cause more information redundancy. To evaluate the effect of $r$, we manually set $r=\{2,20,40,200\}$ and record the average MSE and MAE on Weather, Electricity and ETTh2. The settings roughly cover the range of variable numbers of these datasets. Although a larger number of router tokens may intuitively aggregate information from the $K$$V$ matrices more comprehensively, the results in Fig \ref{fig:r} provide limited support for this assumption. For Electricity, the best $r$ is 10 or 20, while $r=2$ yields best performance on ETTh2. This may be due to the potentially sparse dependencies between variables. Specifically, while complex associations may exist among multivariate data, the patterns of the correlations are likely to be similar. Therefore, a small number of routing tokens could also filter out the redundant information, enhancing the model to focus on more critical inter-series dependencies. Nonetheless, the model performance the overall model performance does not exhibit substantial fluctuations with varying $r$, which demonstrates the robustness of our proposed model.

\begin{figure}[t]
    \centering
    \includegraphics[width=1.0\linewidth]{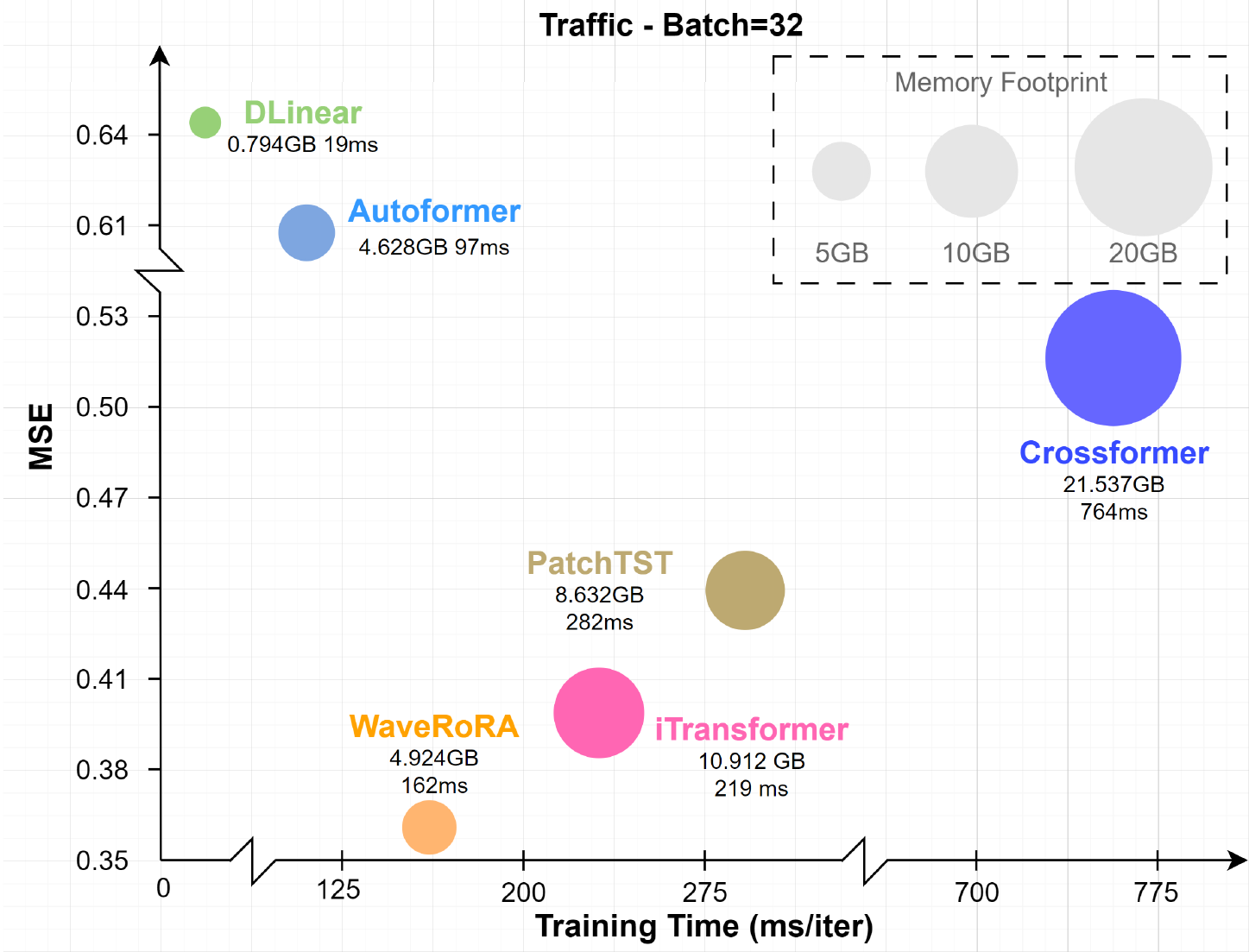}
    \caption{The model efficiency comparison. All the models follow the hyper-parameter settings in section \ref{sec:setting} and the batch size is set to 32.}
    \label{fig:efficiency}
\end{figure}

\subsection{Effect of Varying Wavelet Basis}
Unlike the Fourier transform, the results of wavelet transform vary greatly depending on the chosen wavelet function $\psi$ and scale function $\phi$. Therefore, to explore how different wavelet bases impact model prediction performance across various types of datasets, we selected three wavelet bases: Haar wavelet, Symlet wavelet and Coiflet wavelet\cite{daubechies1992ten}. The Haar wavelet, one of the earliest orthogonal wavelet functions with compact support, is the simplest wavelet function used in wavelet analysis. While computationally efficient, it has limited performance as a wavelet basis due to its discontinuity in the time domain. The Symlet wavelet is a nearly symmetric wavelet function that improves the Daubechies function. It reduces phase distortion to some extent during signal analysis and reconstruction. Generally, the Symlet wavelet is more beneficial for signals with stronger periodic characteristics. In this paper, we primarily use the Sym3 wavelet for DWT. The Coiflet wavelet, which is known for its higher symmetry, could simultaneously maintain high vanishing moments on its wavelet function and scale function. It usually performs well in handling signals with abrupt features. In this experiment, we use Coif3 to evaluate the effectiveness of Coiflet wavelet. The results can be seen in Fig \ref{fig:basis}. For Traffic and Electricity that exhibit more pronounced periodic characteristics, the Sym3 wavelet leads to the best predicting performance, while for ETTh1, the best wavelet function changes to Coif3. This indicates that different wavelet functions can provide tokens with varying quality when handling data with different characteristics.

\subsection{Model Efficiency}
As discussed in section \ref{sec:rora}, RoRA has linear complexity $O(MrD)$ relative to the variable number and the token dimension. To visually evaluate the efficiency of WaveRoRA and other baseline models, we set $r=20$ and batch size $B=32$ on Traffic and record the training speed and GPU memory usage for WaveRoRA, iTransformer, PatchTST, Autoformer, Crossformer and DLinear. The results are shown in Fig \ref{fig:efficiency}. WaveRoRA achieves a good balance among predicting performance, training speed and memory usage.

% \subsection{Visualization}
\section{Conclusion}
In this paper, we investigate to extract the underlying patterns of MTS data in the wavelet domain. We propose a wavelet learning framework to map the input MTS to wavelet embeddings and capture the inter-series dependencies with a novelly designed RoRA mechanism. The embeddings in the wavelet domain could leverage the strengths of both time and frequency domains, modeling the trends and periodic changing patterns of the time series. What's more, the proposed RoRA introduces a small number of routing tokens to aggregate information from the $KV$ matrices and redistribute it to the $Q$ matrix. Such designs retain the powerful expressivity of Softmax and achieve linear complexity relative to both sequence length and token dimension. We conduct extensive experiments and WaveRoRA achieves superior performance compared to SOTA methods. Currently, our wavelet learning framework simply fits for models capturing inter-series dependencies. What's more, different types of wavelet functions exhibit potentials for processing MTS data with varying patterns. Future researches will be done to (a) explore the combination of wavelet domain features with other types of backbones and (b) find an automatic approach to adaptively select the best wavelet function for specific data. In addition, we are going to try applying WaveRoRA on more kinds of applications such as Large Time Series Model\cite{liu2024timer} and AIOPS\cite{diaz2023joint}.

% \bibliography{main}

\begin{thebibliography}{10}

\bibitem{zhang2022solar}
G.~Zhang, D.~Yang, G.~Galanis, and E.~Androulakis, ``Solar forecasting with hourly updated numerical weather prediction,'' {\em Renewable and Sustainable Energy Reviews}, vol.~154, p.~111768, 2022.

\bibitem{uremovic2022new}
N.~Uremovi{\'c}, M.~Bizjak, P.~Suki{\v{c}}, G.~{\v{S}}tumberger, B.~{\v{Z}}alik, and N.~Luka{\v{c}}, ``A new framework for multivariate time series forecasting in energy management system,'' {\em IEEE Transactions on Smart Grid}, 2022.

\bibitem{guo2019attention}
S.~Guo, Y.~Lin, N.~Feng, C.~Song, and H.~Wan, ``Attention based spatial-temporal graph convolutional networks for traffic flow forecasting,'' in {\em Proceedings of the AAAI conference on artificial intelligence}, vol.~33, pp.~922--929, 2019.

\bibitem{hewamalage2021recurrent}
H.~Hewamalage, C.~Bergmeir, and K.~Bandara, ``Recurrent neural networks for time series forecasting: Current status and future directions,'' {\em International Journal of Forecasting}, vol.~37, no.~1, pp.~388--427, 2021.

\bibitem{vaswani2017attention}
A.~Vaswani, N.~Shazeer, N.~Parmar, J.~Uszkoreit, L.~Jones, A.~N. Gomez, {\L}.~Kaiser, and I.~Polosukhin, ``Attention is all you need,'' {\em Advances in neural information processing systems}, vol.~30, 2017.

\bibitem{zhou2021informer}
H.~Zhou, S.~Zhang, J.~Peng, S.~Zhang, J.~Li, H.~Xiong, and W.~Zhang, ``Informer: Beyond efficient transformer for long sequence time-series forecasting,'' in {\em Proceedings of the AAAI conference on artificial intelligence}, vol.~35, pp.~11106--11115, 2021.

\bibitem{nie2023patch}
Y.~Nie, N.~H. Nguyen, P.~Sinthong, and J.~Kalagnanam, ``A time series is worth 64 words: Long-term forecasting with transformers,'' in {\em The Eleventh International Conference on Learning Representations, {ICLR} 2023, Kigali, Rwanda, May 1-5, 2023}, OpenReview.net, 2023.

\bibitem{torres2021deep}
J.~F. Torres, D.~Hadjout, A.~Sebaa, F.~Mart{\'\i}nez-{\'A}lvarez, and A.~Troncoso, ``Deep learning for time series forecasting: a survey,'' {\em Big Data}, vol.~9, no.~1, pp.~3--21, 2021.

\bibitem{wu2023timesnet}
H.~Wu, T.~Hu, Y.~Liu, H.~Zhou, J.~Wang, and M.~Long, ``Timesnet: Temporal 2d-variation modeling for general time series analysis,'' in {\em International Conference on Learning Representations}, 2023.

\bibitem{liu2022msdr}
D.~Liu, J.~Wang, S.~Shang, and P.~Han, ``Msdr: Multi-step dependency relation networks for spatial temporal forecasting,'' in {\em Proceedings of the 28th ACM SIGKDD conference on knowledge discovery and data mining}, pp.~1042--1050, 2022.

\bibitem{zhang2024sageformer}
Z.~Zhang, L.~Meng, and Y.~Gu, ``Sageformer: Series-aware framework for long-term multivariate time series forecasting,'' {\em IEEE Internet of Things Journal}, 2024.

\bibitem{liu2023itransformer}
Y.~Liu, T.~Hu, H.~Zhang, H.~Wu, S.~Wang, L.~Ma, and M.~Long, ``itransformer: Inverted transformers are effective for time series forecasting,'' in {\em The Twelfth International Conference on Learning Representations}, 2023.

\bibitem{jia2024witran}
Y.~Jia, Y.~Lin, X.~Hao, Y.~Lin, S.~Guo, and H.~Wan, ``Witran: Water-wave information transmission and recurrent acceleration network for long-range time series forecasting,'' {\em Advances in Neural Information Processing Systems}, vol.~36, 2024.

\bibitem{zhang2023crossformer}
Y.~Zhang and J.~Yan, ``Crossformer: Transformer utilizing cross-dimension dependency for multivariate time series forecasting,'' in {\em The eleventh international conference on learning representations}, 2023.

\bibitem{cooley1965algorithm}
J.~W. Cooley and J.~W. Tukey, ``An algorithm for the machine calculation of complex fourier series,'' {\em Mathematics of computation}, vol.~19, no.~90, pp.~297--301, 1965.

\bibitem{liu2023temporal}
Z.~Liu, Q.~Ma, P.~Ma, and L.~Wang, ``Temporal-frequency co-training for time series semi-supervised learning,'' in {\em Proceedings of the AAAI Conference on Artificial Intelligence}, vol.~37, pp.~8923--8931, 2023.

\bibitem{gibbs1899fourier}
J.~W. Gibbs, ``Fourier's series,'' {\em Nature}, vol.~59, no.~1539, pp.~606--606, 1899.

\bibitem{wu2021autoformer}
H.~Wu, J.~Xu, J.~Wang, and M.~Long, ``Autoformer: Decomposition transformers with auto-correlation for long-term series forecasting,'' {\em Advances in neural information processing systems}, vol.~34, pp.~22419--22430, 2021.

\bibitem{zhou2022fedformer}
T.~Zhou, Z.~Ma, Q.~Wen, X.~Wang, L.~Sun, and R.~Jin, ``Fedformer: Frequency enhanced decomposed transformer for long-term series forecasting,'' in {\em International conference on machine learning}, pp.~27268--27286, PMLR, 2022.

\bibitem{yang2023waveform}
F.~Yang, X.~Li, M.~Wang, H.~Zang, W.~Pang, and M.~Wang, ``Waveform: Graph enhanced wavelet learning for long sequence forecasting of multivariate time series,'' in {\em Proceedings of the AAAI Conference on Artificial Intelligence}, vol.~37, pp.~10754--10761, 2023.

\bibitem{li2022short}
M.~Li, Y.~Liu, S.~Zhi, T.~Wang, and F.~Chu, ``Short-time fourier transform using odd symmetric window function,'' {\em Journal of Dynamics, Monitoring and Diagnostics}, vol.~1, no.~1, pp.~37--45, 2022.

\bibitem{daubechies1992ten}
I.~Daubechies, ``Ten lectures on wavelets,'' {\em Society for industrial and applied mathematics}, 1992.

\bibitem{daubechies1990wavelet}
I.~Daubechies, ``The wavelet transform, time-frequency localization and signal analysis,'' {\em IEEE transactions on information theory}, vol.~36, no.~5, pp.~961--1005, 1990.

\bibitem{cotter_2019}
F.~Cotter, {\em Uses of Complex Wavelets in Deep Convolutional Neural Networks}.
\newblock PhD thesis, Apollo - University of Cambridge Repository, 2019.

\bibitem{chen2023long}
Z.~Chen, M.~Ma, T.~Li, H.~Wang, and C.~Li, ``Long sequence time-series forecasting with deep learning: A survey,'' {\em Information Fusion}, vol.~97, p.~101819, 2023.

\bibitem{li2019enhancing}
S.~Li, X.~Jin, Y.~Xuan, X.~Zhou, W.~Chen, Y.-X. Wang, and X.~Yan, ``Enhancing the locality and breaking the memory bottleneck of transformer on time series forecasting,'' {\em Advances in neural information processing systems}, vol.~32, 2019.

\bibitem{bai2023crossfun}
Y.~Bai, J.~Wang, X.~Zhang, X.~Miao, and Y.~Lin, ``Crossfun: Multi-view joint cross fusion network for time series anomaly detection,'' {\em IEEE Transactions on Instrumentation and Measurement}, 2023.

\bibitem{zhang2022first}
X.~Zhang, X.~Jin, K.~Gopalswamy, G.~Gupta, Y.~Park, X.~Shi, H.~Wang, D.~C. Maddix, and Y.~Wang, ``First de-trend then attend: Rethinking attention for time-series forecasting,'' {\em arXiv preprint arXiv:2212.08151}, 2022.

\bibitem{wu2020connecting}
Z.~Wu, S.~Pan, G.~Long, J.~Jiang, X.~Chang, and C.~Zhang, ``Connecting the dots: Multivariate time series forecasting with graph neural networks,'' in {\em Proceedings of the 26th ACM SIGKDD international conference on knowledge discovery \& data mining}, pp.~753--763, 2020.

\bibitem{jiang2023fecam}
M.~Jiang, P.~Zeng, K.~Wang, H.~Liu, W.~Chen, and H.~Liu, ``Fecam: Frequency enhanced channel attention mechanism for time series forecasting,'' {\em Advanced Engineering Informatics}, vol.~58, p.~102158, 2023.

\bibitem{wan2024tcdformer}
J.~Wan, N.~Xia, Y.~Yin, X.~Pan, J.~Hu, and J.~Yi, ``Tcdformer: A transformer framework for non-stationary time series forecasting based on trend and change-point detection,'' {\em Neural Networks}, vol.~173, p.~106196, 2024.

\bibitem{katharopoulos2020transformers}
A.~Katharopoulos, A.~Vyas, N.~Pappas, and F.~Fleuret, ``Transformers are rnns: Fast autoregressive transformers with linear attention,'' in {\em International conference on machine learning}, pp.~5156--5165, PMLR, 2020.

\bibitem{liu2022non}
Y.~Liu, H.~Wu, J.~Wang, and M.~Long, ``Non-stationary transformers: Exploring the stationarity in time series forecasting,'' {\em Advances in Neural Information Processing Systems}, vol.~35, pp.~9881--9893, 2022.

\bibitem{kim2021reversible}
T.~Kim, J.~Kim, Y.~Tae, C.~Park, J.-H. Choi, and J.~Choo, ``Reversible instance normalization for accurate time-series forecasting against distribution shift,'' in {\em International Conference on Learning Representations}, 2021.

\bibitem{su2024roformer}
J.~Su, M.~Ahmed, Y.~Lu, S.~Pan, W.~Bo, and Y.~Liu, ``Roformer: Enhanced transformer with rotary position embedding,'' {\em Neurocomputing}, vol.~568, p.~127063, 2024.

\bibitem{zeng2023transformers}
A.~Zeng, M.~Chen, L.~Zhang, and Q.~Xu, ``Are transformers effective for time series forecasting?,'' in {\em Proceedings of the AAAI conference on artificial intelligence}, vol.~37, pp.~11121--11128, 2023.

\bibitem{elfwing2018sigmoid}
S.~Elfwing, E.~Uchibe, and K.~Doya, ``Sigmoid-weighted linear units for neural network function approximation in reinforcement learning,'' {\em Neural networks}, vol.~107, pp.~3--11, 2018.

\bibitem{diaz2023joint}
J.~Diaz-De-Arcaya, A.~I. Torre-Bastida, G.~Z{\'a}rate, R.~Mi{\~n}{\'o}n, and A.~Almeida, ``A joint study of the challenges, opportunities, and roadmap of mlops and aiops: A systematic survey,'' {\em ACM Computing Surveys}, vol.~56, no.~4, pp.~1--30, 2023.

\bibitem{han2023flatten}
D.~Han, X.~Pan, Y.~Han, S.~Song, and G.~Huang, ``Flatten transformer: Vision transformer using focused linear attention,'' in {\em Proceedings of the IEEE/CVF international conference on computer vision}, pp.~5961--5971, 2023.

\bibitem{liu2024timer}
Y.~Liu, H.~Zhang, C.~Li, X.~Huang, J.~Wang, and M.~Long, ``Timer: Transformers for time series analysis at scale,'' {\em arXiv preprint arXiv:2402.02368}, 2024.

\end{thebibliography}
% \bibliographystyle{ieeetr}

\end{document}